\documentclass{egpubl}

 \JournalSubmission    
 \electronicVersion 

\ifpdf \usepackage[pdftex]{graphicx} \pdfcompresslevel=9
\else \usepackage[dvips]{graphicx} \fi

\PrintedOrElectronic

\usepackage{t1enc,dfadobe}

\usepackage{egweblnk}
\usepackage{cite}
\let\rm=\rmfamily        \let\tt=\ttfamily
\let\it=\itshape          
\let\bf=\bfseries

\usepackage{color}
\usepackage{algorithm}
\usepackage{algorithmicx}
\usepackage{scrextend}

\newcommand{\smooth}[1]{\hat{#1}}
\newcommand{\TVL}{$\mbox{\textsc{TV-L}}_{1}$}

\newcommand{\getAlbedoTh}{getAlbedoTh}

\newcommand{\getDepthCues}{getOcclusionGradient}

\definecolor{maroon}{rgb}{1.0,0.5,0}

\definecolor{green}{rgb}{0.8,0,0}

\definecolor{orange}{rgb}{0,0.8,0}

\definecolor{reviewer}{rgb}{1.0,0.0,1.0}
\definecolor{change}{rgb}{0.0,0.0,0.7}
\newcommand{\changed}[1]{#1}

\definecolor{remove}{rgb}{0.7,0.7,0.7}
\newcommand{\removed}[1]{}

\title[Intrinsic Light Field Images]%
      {Intrinsic Light Field Images}

\author[E. Garces, J.I. Echevarria, W. Zhang, H. Wu, K. Zhou \& D. Gutierrez]
       {Elena Garces$^{1}$ \qquad Jose I. Echevarria$^{1,3}$ \qquad Wen Zhang$^{2}$ \qquad Hongzhi Wu$^{2}$ \qquad Kun Zhou$^{2}$ \qquad Diego Gutierrez$^{1}$
        \\
         $^1$ Universidad de Zaragoza, I3A \qquad $^2$ State Key Lab of CAD \& CG, Zhejiang University \qquad $^3$ Adobe Systems
       }

\begin{document}


\maketitle

\begin{abstract}
We present a method  to automatically decompose a \textit{light field} into its intrinsic shading and albedo components. Contrary to previous work targeted to 2D single images and videos, a light field is a 4D structure that captures non-integrated incoming radiance over a discrete angular domain. This higher dimensionality of the problem renders previous state-of-the-art algorithms impractical either due to their cost of processing a single 2D slice, or their inability to enforce proper coherence in additional dimensions. We propose a new decomposition algorithm that jointly optimizes the whole light field data for proper angular coherence. For efficiency, we extend Retinex theory, working on the gradient domain, where new albedo and occlusion terms are introduced. Results show our method provides 4D intrinsic decompositions difficult to achieve with previous state-of-the-art algorithms.
We further provide a comprehensive analysis and comparisons with existing intrinsic image/video decomposition methods on light field images.


\end{abstract}

\section{Introduction}

Intrinsic scene decomposition is the problem of separating the integrated radiance from a captured scene, into physically-based and more meaningful reflectance and shading components, so that $Scene=Albedo \times Shading$; enabling quick and intuitive edits of the materials or lighting in a scene.

However, this decomposition is a very challenging, ill-posed problem. Given the interplay between the illumination, geometry and materials of the scene, there are more unknowns than equations for each pixel of the captured scene. To mitigate this uncertainty, existing \textit{intrinsic decomposition} methods assume that some additional properties of the scene are known. However, the prevailing goal is always the same: the  gradients of the depicted scene need to be classified as coming from a variation in albedo, shading, or both. In this work, we build on classical Retinex theories to obtain better predictors of these variations leveraging 4D information from the light field data.

At the same time, light field photography is becoming more popular, as multi-view capabilities are progressively introduced in commercial cameras~\cite{Lytro,Raytrix}, including mobile devices~\cite{Pelican}. Such captured light fields are 4D structures that store both spatial and angular information of the radiance that reaches the sensor of the camera. This means a correct intrinsic decomposition has to be coherent in the angular domain, which increases the complexity with respect to 2D single images and 3D videos ($x,y,t$). Not only because of the number of additional information to be processed, but also because of the kind of coherence required.

A na\"{\i}ve approach to intrinsic light field decomposition would be to apply any state-of-the-art single image algorithm to each view of the light field independently. However, apart from not taking advantage of the additional information provided by multiple views, angular coherence is not guaranteed. Hence, additional processing would be required to make all the partial solutions, typically around $9\times9$, converge into a single one.
Another approach could be to extend intrinsic video decompositions to 4D light field volumes, as these techniques rely on providing an initial solution for a 2D frame (usually the first), which is then propagated along the temporal dimension. \changed{These algorithms are already designed to keep consistence between frames, but they do not respect the 4D structure in a light field as all images need to be arranged as a single sequence, where the optimal arrangement is unknown. Moreover, the 2D nature of the decomposition propagated back and forth does not fully exploit the information implicitly captured in 4D.}

Therefore, we propose an approach that jointly optimizes for the whole light field data, leveraging its structure for better cues and constraints for solving the problem; and enforcing proper angular coherence by design.
We test our algorithm on both synthetic light fields, and real world ones captured with Lytro cameras. Our results demonstrate the benefits of working in 4D in terms of coherence and quality of the decomposition itself.

\section{Related Work}

Intrinsic decomposition of the shading and albedo components of an image is a long-standing problem in computer vision and graphics since it was formulated by Barrow and Tenembaum in the 70s~\cite{Barrow1972}.  We review previous intrinsic decomposition algorithms based on their input, and then briefly cover related light field processing.

\paragraph*{Single Image.}
Several works rely on the original Retinex theory~\cite{Land71} to estimate the \emph{shading} component. By assuming
that shading varies smoothly, either pixel-wise \cite{tappen2005recovering,zhao2012closed} or cluster-based~\cite{garces2012intrinsic} optimization is performed. Clustering strategies have also been used to obtain the \emph{reflectance} component, e.g. assuming a sparse number of reflectances \cite{Gehler2011,LiShen2011}, using a dictionary of learned reflectances from crowd-sourced experiments~\cite{Bell2014}, or flattening the image to remove shading variations~\cite{Bi2015}. Alternative methods require user interaction~\cite{bousseau2009user}, jointly optimize the shape, albedo and illumination~\cite{Barron2015}, incorporate priors from data driven statistics~\cite{Zhou2015}, train a Convolutional Neural Network (CNN) with synthetic datasets~\cite{Narihira2015}, or use depth maps acquired with a depth camera to help disambiguate shading from reflectance~\cite{Barron2013,Chen2013,Lee:2012}. For a full review of single image methods, we refer the reader to the state-of-the-art~\cite{BKPB17}. \changed{Although some of these algorithms can produce good quality results, they require additional processing for angular coherence, and they do not make use of the implicit information captured by a light field. Our work is based on the Retinex theory, with 2D and 4D scene-based heuristics to classify reflectance gradients.}

\paragraph*{Multiple Images and Video.}
Several works leverage information from multiple images of the same scene from a fixed viewpoint under varying illumination \cite{weiss2001deriving,Hauagge2014,Laffont2015,sunkavalli2007factored}. Laffont et al.~\cite{Laffont2012} coarsely estimate a 3D point cloud of the scene from non-structured image collections. Pixels with similar chromaticity and orientation in the point cloud will be used as reflectance constraints within an optimization.
Assuming outdoor environments, the work of Duchene et al.~\cite{Duchene2015} estimates sunlight position and orientation and reconstructs a 3D model of the scene, taking as input several captures of the same scene under constant illumination. Although a light field can be seen as a structured collection of images, \changed{we do not make assumptions about the lighting nor the scale of the captured scene.}

\paragraph*{Video.}
A few methods dealing with intrinsic video have been recently presented. Ye et al.~\cite{Ye2014} propose a probabilistic solution based a casual-anticasual, coarse-to-fine iterative reflectance propagation. Bonneel et al.~\cite{Bonneel2014} present an efficient gradient-based solver which allows interactive decompositions. Kong et al.~\cite{Kong2014} rely on optical flow to estimate surface boundaries to guide the decomposition. Recently, Meka et al.~\cite{meka:2016} presented a novel variational approach suitable for real-time processing, based on a hierarchical coarse-to-fine optimization. \changed{While this approach can provide coherent and stable results even applied straightforwardly to light fields,
the actual decomposition is performed on a per-frame basis, so it shares the limitations with previous 2D methods.}

\paragraph*{Light fields.} \changed{Related work on intrinsic decomposition of light field images and videos has been published concurrently. Bonneel et al.~\cite{Boneel2017} present a general approach for stabilizing the results of \emph{per-frame} image processing algorithms over an array of images and videos. Their approach can produce very stable results, but its generality does not exploit a 4D structure that can be used to handle complex non-lambertian materials~\cite{Tao2015,Sulc2016}. On the other hand, Alperovich and Goldluecke~\cite{Alperovich2016} present an approach similar to ours posing the problem in ray space. By doing this, they ensure angular coherence and also handle non-lambertian materials. While we do not handle such materials explicitly, our algorithm produces sharper and more stable results, with comparable reconstructions of reflectances under specular highlights.}

\paragraph*{Light Field Editing.} Our work is also related to papers that extend common tools and operations for 2D images to 4D light fields. This is not a trivial task, again given the higher dimensionality of light fields. Jarabo et al.~\cite{Jarabo:SIG14} present a first study to evaluate different light field editing interfaces, tools and workflows, this study is further analyzed by Masia et al.~\cite{Masia2014lf}, providing a detailed description of subjects' performance and preferences for a number of different editing tasks.
Global propagation of user strokes has also been proposed, using a voxel-based representation~\cite{Seitz2002}, a multi-dimensional downsampling approach~\cite{Jarabo2011lf}, or preserving view coherence by reparameterizing the light field~\cite{Ao2015}, while other works focus on deformations and warping of the light field data~\cite{Birklbauer:2016,Chen:2005,Zhang:2002}.
Cho et al.~\cite{Cho2014} utilize the epipolar plane image to extract consistent alpha mattes of a light field. Guo et al.~\cite{Guo2015} stitch multiple light fields via multi-resolution, high dimensional graph cuts. There are also considerable interests in recovering depths from a light field. Existing techniques exploit defocus and correspondence depth cues~\cite{Tao2013}, carefully handle occlusions~\cite{Wang2015}, or use variational methods~\cite{Wanner2014}. As most of these works, we also rely on the epipolar plane for implicit multi-view correspondences and processing.

\begin{figure*}[htb]
\centering
\includegraphics[width=1\textwidth]{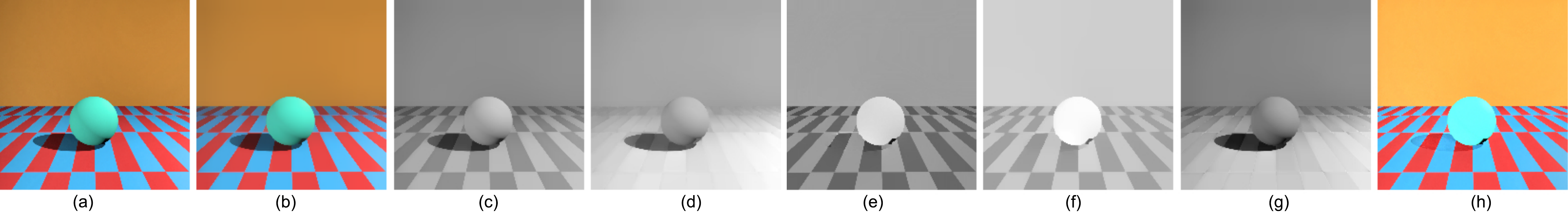}
\caption{Complete pipeline with a simple scene~\protect\cite{Ao2015}. The central view is shown here and the whole light field is shown in the Supplementary Material~\protect\cite{GarcesSuppLF}. (a) Input light field $L$. (b) Filtered light field $\smooth{L}$. (c) Normalized input $||\smooth{L}||_2$. (d) Resulting shading $S_1$ from line 10 in \ref{algorithm:intrinsiclf} and Equation~\ref{eq:lfretinex}; note that although it looks consistent in one view, the global coherence is not guaranteed as shown in the Supplementary Material videos. (e) Resulting reflectance $R_{1}$ from from line 10 in \ref{algorithm:intrinsiclf} and Equation~\ref{eq:lfretinex}. (f) Filtered reflectance $\smooth{R}_{1}$. (g) Final shading $S_f$. (h) Final reflectance $R_f$.}
\label{fig:pipeline}
\end{figure*}

\section{Formulation}

To represent a light field, we use the two-plane parametrization on ray space $L(x,y,u,v)$, which captures a light ray passing through two parallel planes: the sensor plane $\Pi_{uv}$, and the virtual camera plane or image plane $\Omega_{xy}$. Analogous to its 2D image counterpart, the problem of intrinsic light field decomposition can be formulated as follows: for each ray of the light field $L$, we aim to find its corresponding reflectance and shading components $R$ and $S$, respectively.
\begin{equation}\label{eq:intrinsic}
L(x,y,u,v) = R(x,y,u,v) \times S(x,y,u,v)
\end{equation}
Instead of solving for single rays directly, the problem can be formulated in the gradient domain for the image plane $\Omega_{xy}$:
\begin{equation}\label{eq:grad}
\nabla l(x,y,u^*,v^*) = \nabla r(x,y,u^*,v^*) + \nabla s(x,y,u^*,v^*)
\end{equation}
more compactly $\nabla l = \nabla r + \nabla s$. Where $l$, $r$ and $s$ denote the single views for each $\{u^*,v^*\} \in \Pi_{uv}$ for each input view $l$, its reflectance $r$ and shading $s$ in log spaces. Note that we denote single views computed in log domain with lowercase, while uppercase letters denote the whole light field in the original domain.

The classic Retinex approach~\cite{Land71} proposes a solution to this formulation by classifying each gradient as either shading or albedo. As seen before, different heuristics have been proposed over the years, with the simplest one associating changes in albedo with changes in \emph{chromaticity}. Although this provides compelling results for some scenes, it still has the following limitations: chromatic changes do not always correspond to albedo changes; the solution is very sensitive to high frequency texture; and more importantly it does not take into account the effects of occlusion boundaries, where shading and albedo vary at the same time.

\section{Our method}

\subsection{Overview}

Our approach to the problem of intrinsic light field decomposition is based on a multi-level solution detailed in Algorithm~\ref{algorithm:intrinsiclf}:
In a first step, we perform a global 4-dimensional $l_1$ filtering operation, which generates a new version of the light field with reduced high frequency textures and noise, promoted relevant gradients and edges, as well as improved angular coherence. The resulting light field, which we call $\smooth{L}$, will serve to initialize a first estimation of the reflectance $R_0$ and shading components $S_0$ (Section~\ref{sec:init}).
These initial estimations will then be used to compute the albedo and occlusion cues needed for the actual intrinsic decomposition, which is done locally per view (Sections~\ref{sec:retinex} and~\ref{sec:global}), benefiting from the previous global processing of the whole light field volume.
A final global 4D $l_1$ filtering operation (Section~\ref{sec:finalopt}) performed over the reflectance  finishes promoting angular coherence and stability, as can be seen in the results section and the Supplementary Material. \changed{The complete pipeline is shown in Figure~\ref{fig:pipeline}.}

\begin{algorithm}[ht]
	\caption{Intrinsic Light Field Decomposition}
 	\label{algorithm:intrinsiclf}
 \begin{algorithmic}[1]

\State \textbf{Input:} Light field $L(x,y,u,v)$
\State \Comment{Initialization (Section~\ref{sec:init})}
\State $\smooth{L} \gets$ \TVL($L$, $\beta=0.05$)
\State $S_{0}$ $\leftarrow$ $||\smooth{L}||_2$
\State $R_{0}$ $\gets$ $\smooth{L} / S_{0}$
    \State\Comment{Global Analysis (Sections~\ref{sec:retinex} and~\ref{sec:global})}
    \State $\omega_a$ $\leftarrow$ \getAlbedoTh({$\smooth{L}$, $R_0$})
    \State $\omega_{occ}$ $\leftarrow$ \getDepthCues({$L_{depth}$})
\State \Comment{Local intrinsic decomposition}
   \State $R_{1}, S_1 \gets $ $\mathcal{G}$($\smooth{L}$, $\omega_a$, $\omega_{occ}$)  \Comment{Note that $R_{1}$ and $S_1$ are both single channel}
\State \Comment{Global coherence (Section~\ref{sec:finalopt})}
\State $\smooth{R_1} \gets$  \TVL($R_{1}$, $\beta=0.05$)
\State $S_f \gets ||\smooth{L}||_2/\smooth{R_1}$
    \State $R_{f} \gets L / S_{f}$
\State \textbf{Result:} $R = R_{f}(x,y,u,v)$, $S = S_{f}(x,y,u,v)$
\end{algorithmic}
\end{algorithm}

\subsection{Initialization}\label{sec:init}

Inspired by the work of Bi et al.~\cite{Bi2015}, we noticed that better predictions of the albedo discontinuities can be done by performing an initial $l_1$ filtering of the light field volume, since it enhances edges and removes noise that could introduce errors in the estimation of gradients. In particular, we regularize the total variation (TV-$l_1$):
\begin{equation}\label{eq:tvl1}
\min_{\smooth{L}} \frac{1}{2} \| \smooth{L} - L \|^2_2 + \beta \| \smooth{L} \|_{1}
\end{equation}
As a result, from the original light field $L$, we obtain a filtered version $\smooth{L}$, close to the original input but with sharper edges due to the use of $l_1$ norm on the second term. Additionally, the use of this norm effectively removes noise while prevents smoothing out other important features. The regularization factor $\beta$ controls the degree of smoothing, where in our experiments $\beta=0.05$.

Working with light fields means that we need to solve this multidimensional total variation problem in $4D$. For efficiency, we use the ADMM solver proposed by Yang et al.~\cite{Yang:2013}. ADMM combines the benefits of augmented Lagrangian and dual decomposition methods.
It decomposes the original large global problem into a set of independent and small problems, which can be solved exactly and efficiently in parallel. Then it coordinates the local solutions to compute the globally optimal solution.

\begin{figure*}[htb]
\centering
\includegraphics[width=1\textwidth]{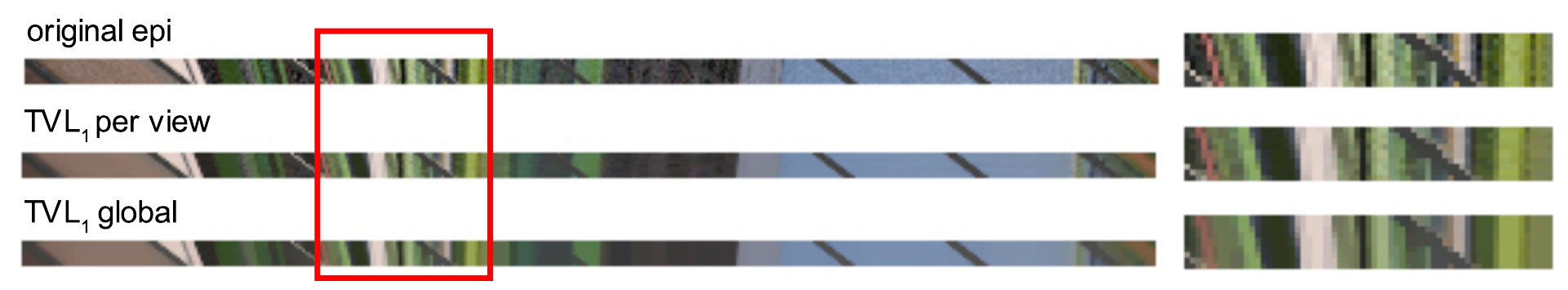}
\vspace{-0.2cm}
\caption{Visualization of the horizonal epi view for the red scanline in Figure~\ref{fig:thcolor}~(a). From top to bottom: the epi from the original light field; the epi after applying $TVL_{1}$ filter to each view separately; the same epi after applying a 4D $TVL_{1}$ filter to the whole light field volume using our approach. We can observe (by zooming in the digital version), areas with very similar colors are flattened, while sharp discontinuities are preserved, effectively removing noise and promoting angular coherence.}
\label{fig:epi}
\end{figure*}

\begin{figure}[htb]
\centering
\includegraphics[width=0.95\columnwidth]{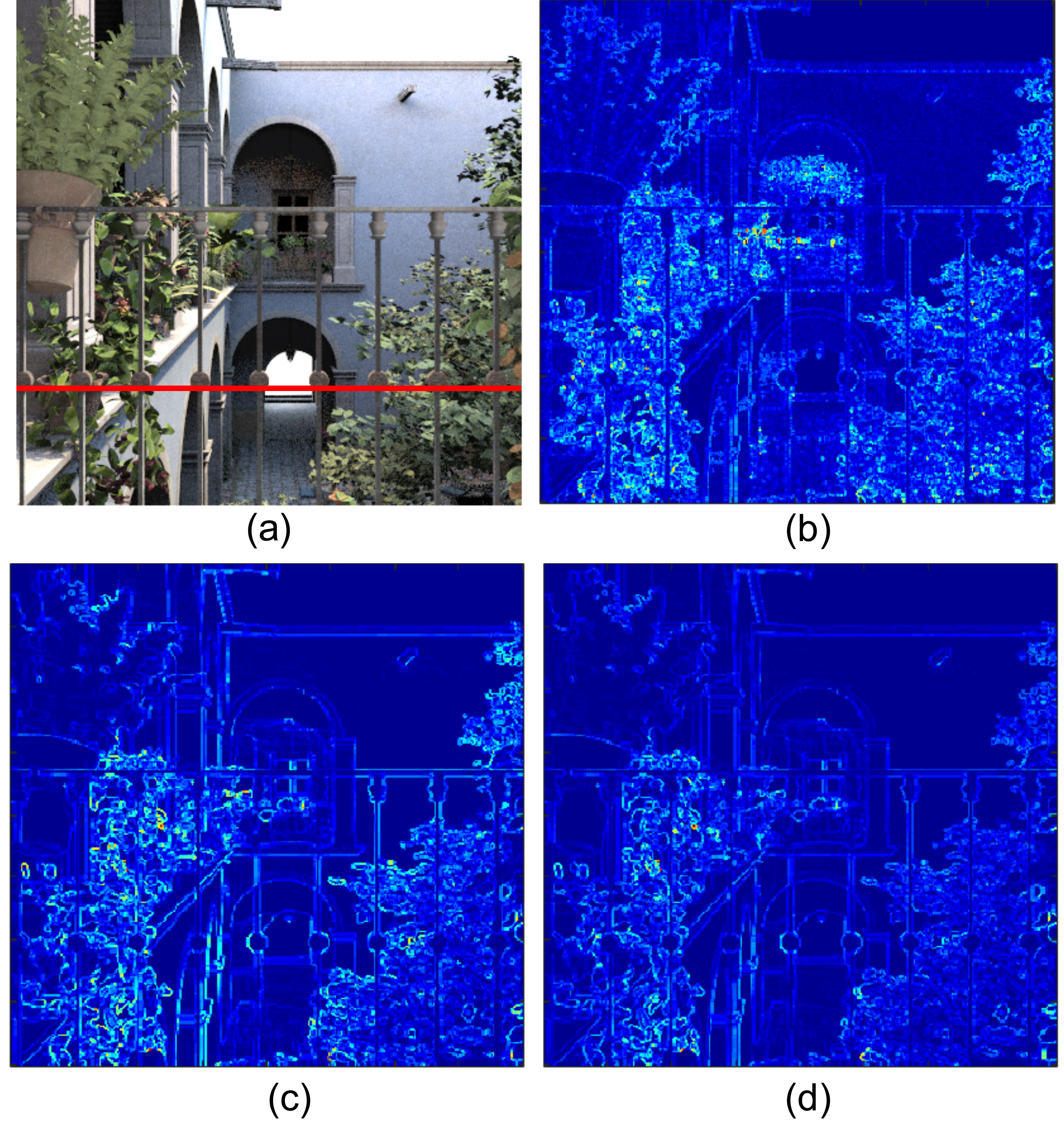}\\
\vspace{-0.2cm}
\caption{(a) Central view of an input light field. (b) Albedo variations computed as the angle between RGB vectors for neighboring pixels $\widehat{L^i, L^j}$, from the original light field $L$. (c) Albedo variations obtained from our initial reflectance estimation, $\widehat{R_0^i, R_0^j}$. (d)~Albedo variation from the chromaticity norm, $||\smooth{L}^i- \smooth{L}^j||$, used by Zhao et al~\protect\cite{zhao2012closed}. Our approach (c) yields cleaner gradients than (b), and captures more subtleties than (d). Note for example the green leaves at the right of the image. Every image is normalized to its maximum value.}
\label{fig:thcolor}
\end{figure}

Figure~\ref{fig:epi} shows the difference in angular coherence and noise between the input $L$, a filtered version obtained from processing each single view independently, and our $\smooth{L}$ obtained from the described global filtering. From $\smooth{L}$, we compute the initial shading as, $S_{0} = ||\smooth{L}||_2$. This is a convenient step to obtain a single-channel version of the input image, with other common transformations like the RGB average or the luminance channel from CIELab~\cite{garces2012intrinsic} providing similar performance.
Taking $S_{0}$ as baseline, we compute the initial RGB reflectance $R_{0}$ simply from $\smooth{L}/S_{0}$. It is important to note that $S_0$ and $R_0$ serve only as the basis over which our heuristics are applied to obtain the final cues to solve for the actual intrinsic decomposition (Equation~\ref{eq:zhao}). Figure~\ref{fig:thcolor} shows the impact of this $l_1$ regularization on the detection of albedo variations.

\subsection{Intrinsic Estimation}

As motivated before, for efficiency, we follow a Retinex approach. We build on Zhao's closed-form formulation~\cite{zhao2012closed}, extending it to take into account our albedo and occlusion cues obtained from the 4D light field volume.
For each view $l$ of the light field,
the system computes the shading component $s$ by minimizing the following equation:
\begin{equation}\label{eq:zhao}
\min_{s} \lambda_1 f_1(s) + \lambda_2 f_2(s) + \lambda_3 f_3(s)
\end{equation}
where $f_1$ is the Retinex constraint, $f_2$ is an absolute scale constraint, and $f_3$ is a non-local texture cue; and $\lambda_1$, $\lambda_2$, and $\lambda_3$ are the weights which control the influence of each term, set to $\lambda_1 = 1$, $\lambda_2 = 1$ and $\lambda_3 = 1000$. In this work we extend $f_1$, so please refer to the original paper for the full details of $f_2$ and $f_3$.

\subsubsection{Retinex-Based Constraint}\label{sec:retinex}

The original Retinex formulation assumes that while shading varies smoothly, reflectance tends to cause sharp discontinuities, which can be expressed as:
\begin{equation}
f_1(s) = \sum_{i,j \in \mathcal{N}_{xy}} (\nabla{s}^2_{ij} + \omega_{ij}^a\nabla{r}_{ij}^2)
\end{equation}
where $\mathcal{N}_{xy}$ is the set of pairs of pixels that can be connected in a four-connected neighborhood defined in the image plane $\Omega_{xy}$,
and $\omega_{ij}^a$ is commonly defined as a threshold on the variations in the chromatic channels (Section~\ref{sec:global}).
Following Equation~\ref{eq:grad}, we define the following transformation, needed to solve Equation~\ref{eq:zhao}.
\begin{equation}\label{eq:transform}
\nabla{r} = \nabla{\smooth{l}} - \nabla{s}
\end{equation}

However, we found that this equation ignores the particular case of occlusion boundaries, where shading and reflectance may vary at the same time. In order to handle such cases, we introduce a new additional term $\omega_{ij}^{occ}$, which has a very low value when an occlusion is detected, so it does not penalize the corresponding gradients (more details in Section~\ref{sec:global}):
\begin{equation}\label{eq:retinex}
f_1(s) = \sum_{i,j \in \mathcal{N}_{xy}}  \omega_{ij}^{occ}(\nabla{s}^2_{ij} + \omega_{ij}^a\nabla{r}_{ij}^2)
\end{equation}

We define $\mathcal{G}$ as the function that takes the whole light field and the global cues to obtain the corresponding shading and reflectance layers:
\begin{equation}
\mathcal{G}(\smooth{L}, \omega^a, \omega^{occ}) = (S_1, R_{1})
\label{eq:lfretinex}
\end{equation}
It is important to note that $s$ has a single channel (an interesting future work would be to lift this restriction to allow colored illumination), so Equation~\ref{eq:transform} is also a single channel operation, where $\smooth{l}$ is $||\smooth{l}||_2$.
Therefore, Equation~\ref{eq:zhao} yields single channel shading $s$, and reflectance $r = ||l||_2 - s$ in log-spaces. Then, $S_1$ and $R_{1}$ are:
\begin{equation}
\forall u,v \in \Pi_{uv} \mbox{     } {\begin{array}{lcl} S_1(x,y,u,v) & = & e^{s} \\
R_{1}(x,y,u,v) & = & e^{r}
\end{array}}
\end{equation}

\subsection{Gradient Labeling}\label{sec:global}
In the following, we describe our extensions to the classic Retinex formulation: the albedo and occlusion terms in Equation~\ref{eq:retinex}. Note that this labeling is independent from solving the actual system (Equation~\ref{eq:zhao}), so each cue is computed in the most suitable color space, or additional available dimensions like depth.

\subsubsection{Albedo Gradient ($\omega^a$)}

Albedo gradients are usually computed based on the chromatic information in CIELab color space. However, as we have shown, our initial RGB reflectance $R_{0}$ is better suited for this purpose, since it shows more relevant albedo variations. \changed{Staying in RGB space, we are inspired by the planar albedo assumption of Bousseau et al.~\cite{bousseau2009user} and propose an edge-based analysis where if neighboring pixels $\{i,j\}$ are co-linear, their albedo is assumed to be constant. This is a heuristic that works reasonably well in practice except for black and white albedo, which are handled separately.}
We thus compute our weights as:

\begin{equation}
\label{eq:weight}
\omega_{ij}^{a} = \left\{ {\begin{array}{*{20}c}
   {0,} & { \mbox{if  } \widehat{R_0^i, R_0^j}} > 0.04 \\
   {1,} & { \mbox{otherwise}}  \\
\end{array}} \right.
\end{equation}
Setting $\omega_{ij}^{a} = 0$ in Equation~\ref{eq:retinex}, means that such gradient comes from albedo, so the gradient of the shading should be smooth. We found a difference of $0.04$ radians works well in general, producing good results. We can see an example in Figure~\ref{fig:thcolor}, where our measure is compared to the original Zhao's estimator, which only used Euclidean distances.

Our proposed heuristic works reasonably well when there is color information available, however it fails when colors are close to pure black or white. Thus, we choose to detect them independently and use them as similar cues as for regular albedo, so the final shading is not affected.
We propose an approach based on the distance from a color to the black and white references in CIELab space (given its better perceptual uniformity than RGB), which gives a measure of the probability of a color being one of them.

\begin{figure*}
\includegraphics[width=1\textwidth]{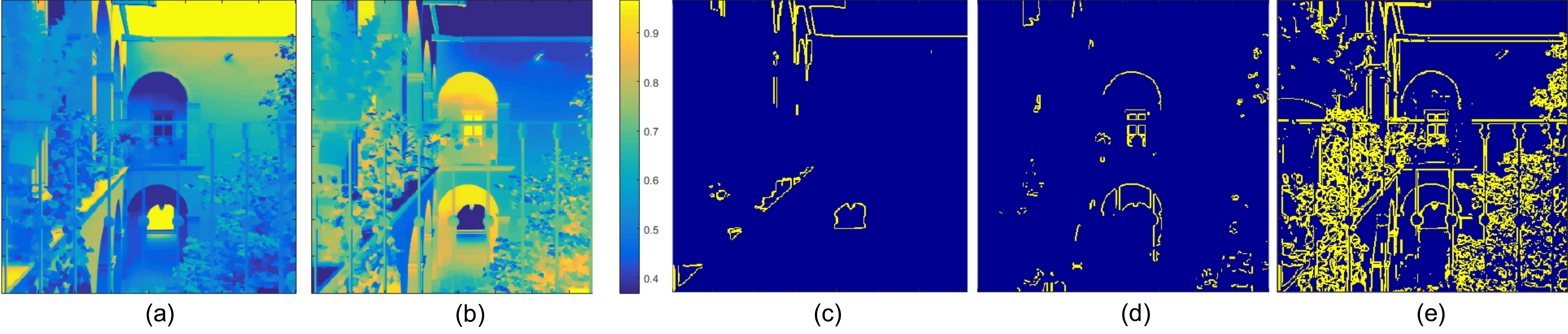}
\vspace{-0.5cm}
\caption{ (a) Probability of being white, $\mathcal{P}_i^w$ (b) Probability of being black, $\mathcal{P}_i^b$ (c) White pixels masked after $g_{ij}^w$ (d) Black pixels masked after $g_{ij}^b$ (e) Final albedo weights $\omega_{ij}^a$ taking into account color, white, and black information.}
\label{fig:th}
\end{figure*}

From the light field $\smooth{L}$, we compute the perceptual distance of each pixel to the white color as $\mathcal{D}_i^w = \| \smooth{L}_i-w \|_2^2$, and analogously the distance to black $\mathcal{D}_i^b$; where $w$ and $b$ may change depending on the implementation. With that, we compute the probability of a pixel of being white or black as $\mathcal{P}_i^w = \exp(-\mathcal{D}_i^w/\mathcal{D}_b^w)$, with $\mathcal{D}_b^w$ being the maximum distance in CIELab space (see Figure~\ref{fig:th}).
Then, we label the gradients as:
\begin{equation}
\label{eq:white}
g_{ij}^{w} = \left\{ {\begin{array}{*{20}c}
   {0,} & { \mbox{if  } (\mathcal{P}_i^w \geq \tau \| \mathcal{P}_j^w \geq \tau_1 ) \wedge (| \mathcal{D}_i^w - \mathcal{D}_j^w | > \tau_2) }\\
   {1,} & { \mbox{otherwise}}  \\
\end{array}} \right.
\end{equation}
where $\tau_1 = 0.85$ and $\tau_2 = 0.05$. And we impose the additional condition that it must be a real gradient, so $| \mathcal{D}_i^w - \mathcal{D}_j^w | > \tau_2$ avoids marking pixels inside uniform areas. The black albedo labeling $g_{ij}^{b}$ is analogously formulated. $\tau_1$  and $\tau_2$ were set empirically, but work well for all tested scenes. Then, we compute the final albedo threshold for each gradient as $\omega_{ij}^a = \max (\omega_{ij}^{a}, g_{ij}^w, g_{ij}^b)$.
The result of this step is a binary labeling, where each gradient is labeled as albedo or shading change (Figure~\ref{fig:th}).

\subsubsection{Occlusion Gradient ($\omega^{occ}$)}
\begin{figure*}
\centering
\includegraphics[width=1\textwidth]{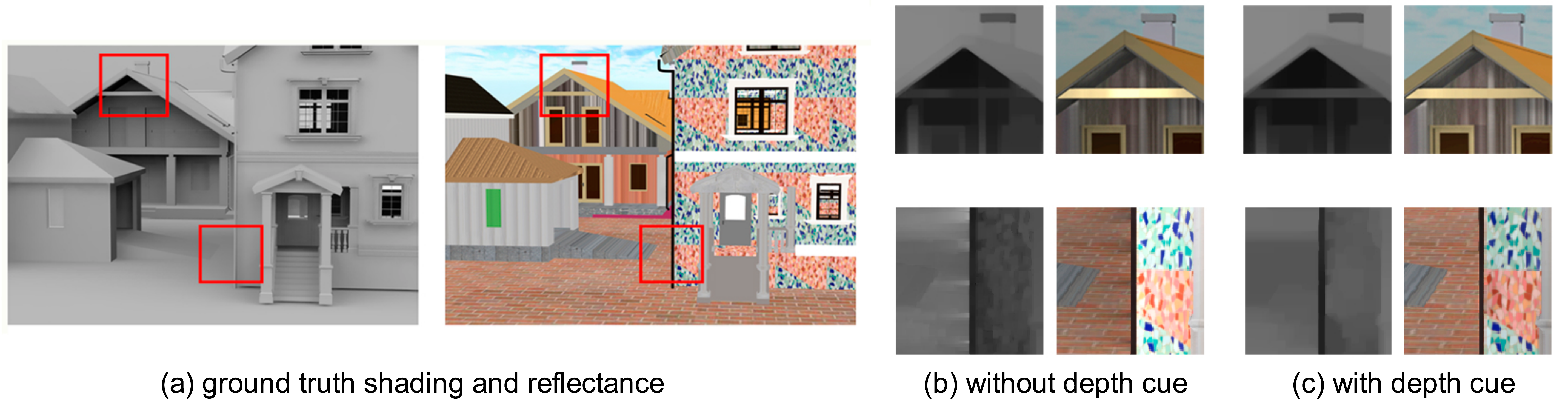}
\vspace{-0.5cm}
\caption{(a) Ground truth shading. (b) Ground truth reflectance. (c) Without $\omega^{occ}$, the algorithm classifies some prominent gradients as albedo, so it enforces continuous shading, causing artifacts. Taking occlusions into account fixes this limitation, producing results closer to the reference.}
\label{fig:threshold}
\end{figure*}

Previous work assume that discontinuities come from changes in albedo or changes in shading, but not both. However, we found they can actually occur simultaneously at occlusion boundaries, becoming an important factor in the intrinsic decomposition problem. Our key idea then is to detect the corresponding gradients and assign them a low weight $\omega_{ij}^{occ}$ in Equation \ref{eq:retinex}, so larger changes are allowed in shading and albedo at the same time.
Contrary to single 2D images, 4D light fields provide several ways to detect occlusions, like analyzing the epipolar planes~\cite{Apostoloff2005,Wanner2014} or using defocus cues~\cite{Wang2015}. In the following, we describe a simple heuristic assuming an available depth map~\cite{Tao2015}, although it can be easily adjusted if only occlusion boundaries are available:

\begin{equation}
\label{eq:white}
\omega_{ij}^{occ} = \left\{ {\begin{array}{*{20}c}
   {0.01,} & { \mbox{if  } | D_i - D_j | > 0.02}\\
   {1,} & { \mbox{otherwise}}  \\
\end{array}} \right.
\end{equation}
where the depth map $D$ is normalized between 0 and 1. Note that we cannot set $\omega_{ij}^{occ} = 0$ because it would cause instabilities in the optimization. Figure~\ref{fig:threshold}~(c), show the effect of including this new term.

\subsection{Global Coherence}\label{sec:finalopt}

After solving Equation~\ref{eq:lfretinex} we get $S_1$ and $R_{1}$. Given the way normalization of shading values is performed in Equation~\ref{eq:zhao}, we found some views may become a bit unstable, affecting the angular coherence of the results. A straightforward approach could be to apply another 4D $l_1$ filter (Equation~\ref{eq:tvl1}) over $S_1$. But, this tends to remove details, wrongly transferring them to the reflectance producing an over-smoothed shading layer and a noisier reflectance one.

We found filtering $R_{1}$ provides better results. Because $R_{1}$ already features uniform regions of color, the 4D $l_1$ filter finishes flattening them for enhanced angular coherence, obtaining $\smooth{R_1}$. Again, we use $\beta = 0.05$.
From there, we compute our final smooth and coherent shading $S_f$ as $||\smooth{L}||_2/\smooth{R_1}$. And the final RGB reflectance as $R_f = L/S_f$.

\section{Results and Evaluation}

We show the whole pipeline in Figure~\ref{fig:pipeline}. The central view is shown after each step of the Algorithm~\ref{algorithm:intrinsiclf}, plus the whole light field is shown in the Supplementary Material~\protect\cite{GarcesSuppLF}. The input light field $L$, the filtered version $\smooth{L}$ and the normalized version $||\smooth{L}_2||$ are shown in Figures~(a) to (c). We observe that the variation between the original light field $L$ and the filtered one $\smooth{L}$ is very subtle. In particular, in this figure, it is more noticeable in very dark regions where black gradients become grayish. This is favorable to the gradient-based solver we use to solve Equation~\ref{eq:zhao}, which is very sensitive to very dark areas \changed{(with values close to zero)}. The output from Equation~\ref{eq:lfretinex} is shown in Figures~(d) and (e), and, although the shading looks pretty consistent in one view, it lacks of angular consistence when the whole volume is visualized (as shown in the Supplementary). Finally, from the filtered reflectance $\smooth{R}_{1}$~(f) and the original light field $L$, we are able to recover the coherent shading $S_f$~(g) and reflectance layers $R_f$~(h). Note that the initial filtering operation also removes small details in shadows and texture, which are recovered in the reflectance layer. This is favorable if the details removed are high frequency texture, as we can see in Figure~\ref{fig:2D}~(first row), but may also cause small remnants of shading in the reflectance, as we can see in Figure~\ref{fig:pipeline}~(h).

\begin{figure*}
\centering
\includegraphics[width=\linewidth]{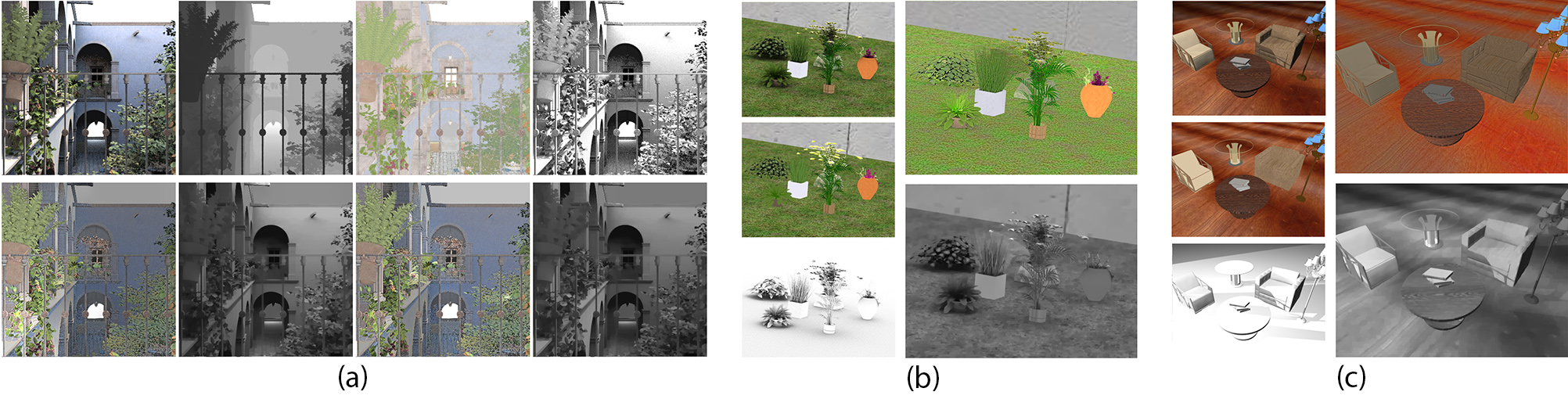}
\vspace{-0.5cm}
\caption{(a) \emph{sanmiguel}. First row: input, depth map and ground truth albedo and shading. Second row: left, our result without occlusion cue; right, our result with occlusion cue. (b) \emph{living room}. Left column: input, ground truth albedo and shading. Right column: our result with occlusion cue. (c) \emph{plants}. Left column: input, ground truth albedo and shading. Right column: our result with occlusion cue.}
\label{fig:sanmiguelfull}
\end{figure*}

\changed{In addition to the scenes shown for the comparisons, we also provide a different set of results with our method in a variety of real and synthetic scenes in our Supplementary Material. In Figure~\ref{fig:sanmiguelfull} we show the full result for \textit{sanmiguel} scene without and with the occlusion cue. In this example, knowing the depth map improves the albedo decomposition as the left-most part of the image is more balanced. In the other two scenes (\textit{plants} and \textit{livingroom}) the difference between both scenarios is more subtle so we just show here the output with the cue. We can observe again that our filtering step favors high frequency albedo details. As has been noted in related work, there is a close relationship between intrinsic estimation and high frequency detail removal~\cite{Bi2015}.}

\begin{figure}
\centering
\includegraphics[width=0.75\columnwidth]{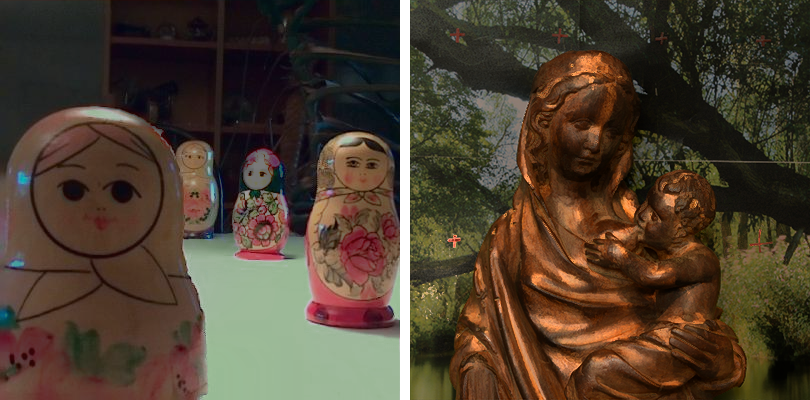}
\caption{Simple editing operations performed by modifying the albedo (left) and shading (right) layers independently. Please check the accompanying videos to see the complete edited light field.}
\vspace{-0.3cm}
\label{fig:edits}
\end{figure}

Intrinsic light field decomposition extends the range of edits that can be performed to a light field with available tools~\cite{Jarabo:SIG14,Masia2014lf}. \changed{Figure~\ref{fig:edits} shows two examples, where simple albedo and shading edits allow to change the appearance coherently across the angular domain. Please note more advanced manipulations like texture replacement are still an open problem in 4D.}

\subsection{Discussion}

\changed{In the following, we discuss and compare our approach with related work and some straightforward alternatives. Our results for the comparisons do not make use of the occlusion cues. For all of them we show the final decomposition for the central view of the light field. Angular coherence can be inspected in the animated sequences included in the Supplementary Material\cite{GarcesSuppLF}.}

\begin{figure*}
\centering
\vspace{1cm}
\includegraphics[width=\textwidth]{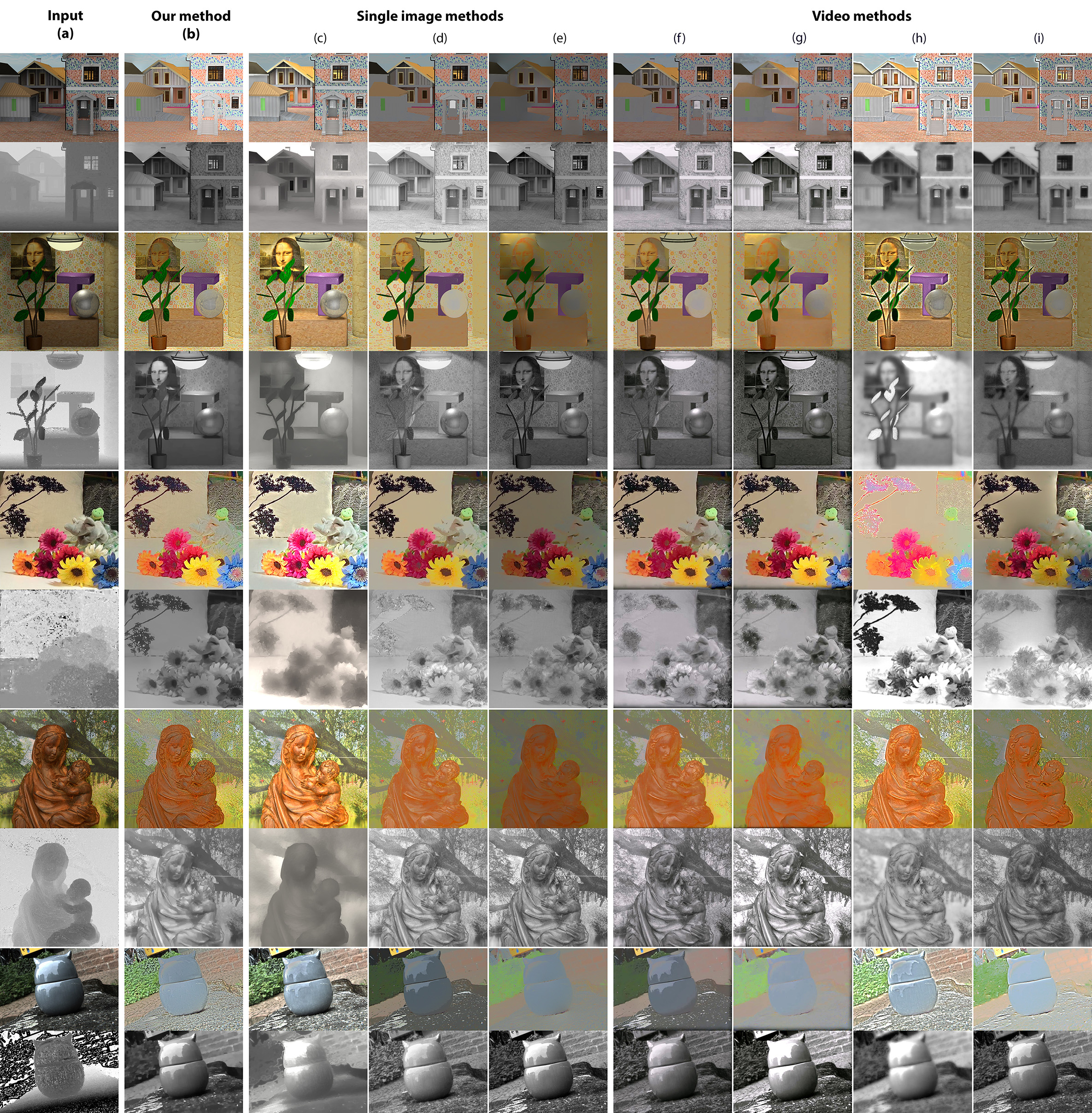}
\caption{\changed{(a) Input RGB and depth data (computed using Wang et al.~\cite{Wang2015}). (b) Our results. Single image approaches: (c) Chen and Koltun~\cite{Chen2013},~(d) Bell et al.~\cite{Bell2014}, (e)~Zhao et al.~\cite{zhao2012closed}. Video approaches: (f)-(g) single image methods ((d) and (e)) filtered using blind temporal consistency~\cite{Boneel2015}, (h) Meka et al.~\cite{meka:2016}, (i) Bonneel et al.~\cite{Bonneel2014}. The scenes are named, from top to bottom: outdoor, Monas' room, frog, Maria, and owlstr.}}
\label{fig:2D}
\end{figure*}

\paragraph*{Single Image.} \changed{Figure~\ref{fig:2D} shows a comparison with 2D state-of-the-art methods that use a single color image as input. The method of Chen et al.~\cite{Chen2013} requires an additional depth map, which in comparable real scenarios could be reconstructed from the light field itself (we use Wang et al.~\cite{Wang2015} for this matter). In terms of overall accuracy of the decomposition, it could be argued that the RGB-D approach provides better results, specially in the shading component. However, results tend to be overly smooth and artifacts appear when the reconstructed depth map is not accurate enough. But more important, this approach requires non-trivial additional processing for solving the remaining views given depth maps are usually computed only for the central view.
Compared to the other single image inputs, our method provides very similar results per view, while it keeps the angular coherence (see the Supplementary Material to observe the flickering artifacts that appear solving the decomposition per view).
Straight processing of the whole array of views as a single image is obviously impractical given the huge number of equations to be solved.}

\begin{figure}
\centering
\includegraphics[width=0.9\columnwidth]{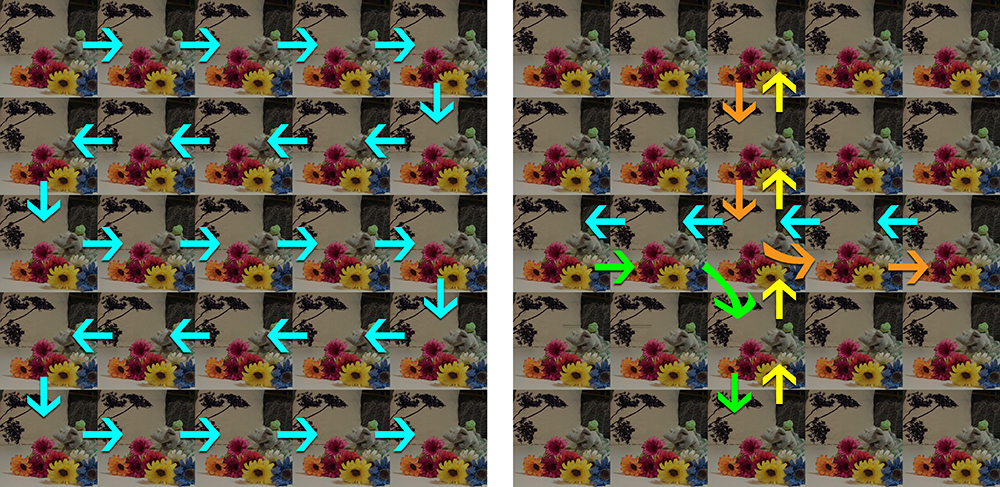}
\caption{\changed{Left: Sequence for video processing. Right: sequence for the animations in the Supplementary Material.}}
\label{fig:seqs}
\end{figure}

\paragraph*{Video.} \changed{If the different views captured in a 4D light field are arranged as a single sequence, they can be interpreted as a video, and so previous intrinsic video solutions can be applied. While the optimal sequence is unknown, we chose the one in Figure~\ref{fig:seqs} (left). Apart from specific intrinsic video algorithms, we also tested a more general approach based on blind temporal consistency~\cite{Boneel2015}, where the single image solutions from the previous paragraph were applied per frame, to be then processed for enhanced coherence (an approach that can be also found in concurrent work~\cite{Boneel2017}). As can be seen in Figure~\ref{fig:2D}, both methods, Bonneel et al.~\cite{Bonneel2014} and Meka et al.~\cite{meka:2016}, produce results that tend to be too smooth, with visible flickering and haloing artifacts when played in a different order from the original sequence (proper angular coherence needs to be independent of the order of visualization). Blind temporal consistency~\cite{Boneel2015} applied over single frames from Zhao et al.~\cite{zhao2012closed} and Bell et al.~\cite{Bell2014} is able to produce stable results when the baseline between views is very little as the per view decompositions are very similar. However, while this seems to be an effective way of enforcing angular coherence, working independently over single frames has some limitations when it comes to extensions to handle non-lambertian surfaces. This is something out of the scope of our paper, but an interesting venue for future work as already demonstrated in related work~\cite{Tao2015,Alperovich2016,Sulc2016}.}

\begin{figure}
\centering
\includegraphics[width=\columnwidth]{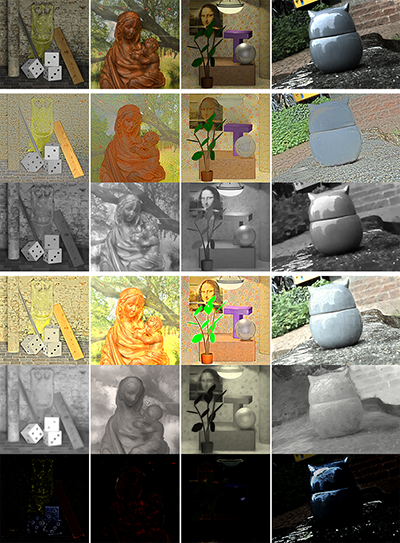}
\caption{\changed{Light field methods. From top to bottom: center view of input light field; our results; results from Alperovich and Goldluecke~\protect\cite{Alperovich2016}, including their additional specular layer. Given this extra layer, it is easier to compare results based on reflectance alone, where we are able to recover more plausible values in areas covered by strong specular highlights.}}
\label{fig:lightFields}
\end{figure}

\vspace{-0.1cm}
\paragraph*{Light Field Images.} \changed{Finally, concurrent work has appeared also decomposing 4D light field images into their intrinsic components. In their paper, Alperovich and Goldluecke~\cite{Alperovich2016} also pose the problem in the 4D ray space, with the additional goal of separating specular reflections from the albedo and shading. Figure~\ref{fig:lightFields} shows comparisons between the processed central views, while the animated sequences in the Supplementary Material showcase angular coherence. From the static images, similar overall quality is achieved. It is interesting to see, however, that although we do not explicitly process specular highlights, our reflectance layers are able to recover better values in some of these regions (mirror ball in \emph{Mona's room} and the blue owl figurine). From the animated sequences, our results show less flickering and so better angular coherence.
It is worth mentioning that because of the computing requirements, we were not able to get the full decomposed light fields from Alperovich and Goldluecke \cite{Alperovich2016}. Our method, however, has still room for optimization, given each 2D view can be solved in parallel, before and after the 4D operations.}

\vspace{-0.1cm}

\section{Conclusions and Future Work}
We have presented a new method for intrinsic \textit{light field} decomposition, which adds to existing approaches for single images and video, enabling practical and intuitive edits in 4D.
Our method is based on the Retinex formulation, reviewed and extended to take into account the particularities and requirements of 4D light field data. We have shown results with both synthetic and real datasets, which compare favorably against existing state-of-the-art methods.

For our albedo and occlusion cues, we currently rely on simple thresholds. A more sophisticated solution could make use of multidimensional Conditional Random Fields~\cite{jampani:cvpr:2016}. Despite the flexibility of our formulation with respect to depth data, a current limitation is that its quality can directly affect the final results. More sophisticated occlusion heuristics could combine information from the epipolar planes to make this term more robust.

\changed{Finally, to reduce the complexity of the intrinsic decomposition problem, some simplifying assumptions are usually made, with the most relevant ones about the color of the lighting (white light) and the material properties of the objects in the scene (non-specular lambertian surfaces). As we have seen, although some approaches adapted from video processing can arguably match our method in terms of stability and quality of the decomposition, extensions to handle more complex materials and scenes can be posed more naturally and effectively in 4D space, paving the way for interesting future work.}

\paragraph*{Acknowledgements}

We thank the reviewers for their insightful comments, Anna Alperovich for their datasets, Nicolas Bonneel and Abhimitra Meka for kindly providing the necessary comparisons, Adrian Jarabo and Belen Masia for fruitful discussions and synthetic scenes.
This research has been funded by the European Research Council (ERC Consolidator Grant, project Chameleon, ref. 682080), as well as the Spanish Ministry of Economy and Competitiveness (projects TIN2016-78753-P and TIN2016-79710-P). The authors from Zhejiang University were partially supported by the National Key Research \& Development Plan of China (2016YFB1001403), NSFC (No. U1609215) and the Fundamental Research Funds for the Central Universities.

\bibliographystyle{eg-alpha-doi}
\bibliography{paper_intrinsiclf}

\end{document}